\documentclass{article}

\usepackage[preprint,nonatbib]{neurips_2020}

\usepackage[utf8]{inputenc} 
\usepackage[T1]{fontenc}    
\usepackage{amsfonts}       
\usepackage{booktabs}       
\usepackage{hyperref}       
\usepackage{url}            
\usepackage{microtype}      
\usepackage{nicefrac}       

\usepackage{array}
\usepackage{epsfig}
\usepackage{amsmath}
\usepackage{amssymb}
\usepackage{multirow}
\usepackage{subfigure}
\usepackage{threeparttable}
\usepackage[toc,page]{appendix}

\title{Inter-layer Information Similarity Assessment of Deep Neural Networks Via Topological Similarity and Persistence Analysis of Data Neighbour Dynamics
}

\author{Andrew Hryniowski$^{1,2,3}$ and Alexander Wong$^{1,2,3}$\\
			$^{1}$ Vision and Image Processing Research Group, University of Waterloo\\
			$^{2}$ Waterloo Artificial Intelligence Institute\\
			$^{3}$ DarwinAI Corp.\\ 			
			\texttt{$\{$apphryni, a28wong$\}$@uwaterloo.ca}
	}

\begin{document}

\maketitle

\begin{abstract}
The quantitative analysis of information structure through a deep neural network (DNN) can unveil new insights into the theoretical performance of DNN architectures.  Two very promising avenues of research towards quantitative information structure analysis are: 1) layer similarity (LS) strategies focused on the inter-layer feature similarity, and 2) intrinsic dimensionality (ID) strategies focused on layer-wise data dimensionality using pairwise information.  Inspired by both LS and ID strategies for quantitative information structure analysis, we introduce two novel complimentary methods for inter-layer information similarity assessment premised on the interesting idea of studying a data sample's neighbourhood dynamics as it traverses through a DNN. More specifically, we introduce the concept of \textbf{Nearest Neighbour Topological Similarity} (NNTS) for quantifying the information topology similarity between layers of a DNN.  Furthermore, we introduce the concept of \textbf{Nearest Neighbour Topological Persistence} (NNTP) for quantifying the inter-layer persistence of data neighbourhood relationships throughout a DNN. The proposed strategies facilitate the efficient inter-layer information similarity assessment by leveraging only local topological information, and we demonstrate their efficacy in this study by performing analysis on a deep convolutional neural network architecture on image data to study the insights that can be gained with respect to the theoretical performance of a DNN.

\end{abstract}

\section{Introduction}
\label{sec:motivation}

Deep neural networks (DNNs) are functions that map information from one domain to another~\cite{lecun2015deep}. These maps often consist of hundreds of sub-maps in the form of element-wise non-linear functions, matrix multiplications, convolutions, etc.~\cite{lecun2015deep}. Each one of these sub-maps gradually warps the underlying manifold of a dataset. Studying the properties of these sub-maps and the effects on a dataset's manifold through a DNN at a micro and macro level can lead to a better understanding of a DNN's internal workings. Therefore, such a quantitative analysis of information structure through a DNN can unveil new insights into the theoretical performance of DNN architectures.

Two very promising avenues of research towards quantitative information structure analysis are: 1) intrinsic dimensionality (ID)~\cite{facco2017estimating, ansuini2019intrinsic} strategies, and 2) layer similarity (LS) strategies~\cite{gretton2005measuring, hardoon2004canonical, raghu2017svcca, kornblith2019similarity}.  At the micro level, intrinsic dimensionality (ID) strategies allow for approximations of a manifold's dimensionality. Lacking from ID analysis is a notation of distance between layers. Knowing the number of dimensions required to represent a manifold does not illuminate the manifold's internal characteristics, and directly comparing the magnitude of the ID between layers provides limited actionable information. On the macro level, layer similarity (LS) strategies~\cite{gretton2005measuring, hardoon2004canonical, raghu2017svcca, kornblith2019similarity} are designed to compare the similarity of information representations between layers. LS measures work by comparing the features of one layer to all the other features of another layer across a set of input data. As such, measuring how a local region of the dataset manifold changes is not possible.

Inspired by both LS and ID strategies for quantitative information structure analysis, we propose a data-centric approach to study the effects a DNN has on the local topological structure of a dataset's manifold based on data neighbourhood dynamics as it traverses through a DNN. More specifically, we construct nearest neighbour graph (NNG) representations to capture the topological structure of a dataset's information representation for each layer in a DNN. Then we compare each layers' NNG using two novel forms of analysis: 1) \textbf{Nearest Neighbour Topological Similarity} (NNTS) to measure the local topological similarity between layers, and 2) \textbf{Nearest Neighbour Topological Persistence} (NNTP) to investigate inter-layer interacts on a pairwise data sample basis. These two proposed approaches open the door for fine-grained analysis of the complex data dynamics present within a DNN. At a high level these methods compare the first degree relations between dataset samples within a layer to such relations in another layer.

\section{Nearest Neighbour Topological Similarity}
\label{sec:nnnn_similarity}

Below is a brief definition of the nearest neighbour graph (NNG) used within this work to capture properties of a dataset's topological structure. See Appendix~\ref{appendix:knn_graph} for the motivation behind the graph's design choices. Let $\textbf{x}_i \in \textbf{X}$ be a set of input samples, and let $G$ represent a DNN. Let the output of some sub-function $v_v$ for the $\textbf{x}_i$ sample be defined as $\textbf{y}_{vi} = v_{v}(\textbf{x}_i; G_v)$, where $G_v \subseteq G$ contains all required sub-functions, edges, and weights to calculate $\textbf{y}_{vi}$. The main idea behind our approach is to use a graph of neighbours to capture the local structure between samples within a layer. For a given layer $v_v$ with a set of outputs $\textbf{Y}_{v}$, let $H_v = (\textbf{Y}_v, D_v)$ be the graph of neighbours for layer $v_v$, where $\textbf{Y}_{v} = v_{v}(\textbf{X}; G_v)$ are the vertices of the graph, and $D_v$ are the edges between two given samples $\textbf{y}_{vi},\textbf{y}_{vj} \in \textbf{Y}_{v}$. Let $K_{vi} \subseteq \textbf{Y}_{v}$ be an ordered set of nearest neighbours of sample $\textbf{y}_{vi}$. Directed edges are used for NNG construction.

Let $Q(H_{a}, H_{b}) = q_{ab}$ measure the Nearest Neighbour Topological Similarity (NNTS) between layers $v_a$ and $v_b$ where $\textbf{y}_{ai}$ and $\textbf{y}_{bi}$ are sample $\textbf{x}_i$'s representation in layers $v_a$ and $v_b$, respectively. To compare a single sample across layers we propose a sample-wise similarity function $Q_{s}(\textbf{y}_{ai}, \textbf{y}_{bi})$ where $Q(\cdot)$ is a function of all $Q_{s}(\cdot)$. Let $Q(H_{a}, H_{b})$ be defined as
\begin{equation}
\label{eq:qk_init}
    Q(H_{a}, H_{b}) = \frac{1}{n} \sum_i^{n} Q_{s}(\textbf{y}_{ai}, \textbf{y}_{bi})
\end{equation}
Then for a given sample $\textbf{x}_{i}$ for layers $a$ and $b$ we get neighbour sets $K_{ai}$ and $K_{bi}$, respectively, for some given $k$. Let the per-sample inter-layer similarity function be defined as the IOU between layers. Note that this formulation also allows a sample to have different neighbours between layers.
\begin{equation}
   Q_{s}(\textbf{y}_{ai}, \textbf{y}_{bi}) = \frac{|K_{ai} \cap K_{bi}|}{|K_{ai} \cup K_{bi}|}
\end{equation}
$Q(\cdot)$ uses local information through first degree relations of a sample within a layer, and compares samples between layers by comparing the local characteristics of different representations of a sample.

\begin{figure}[t]
    \centering
    \begin{tabular}{cccc}
    \includegraphics[trim={1.5cm 1.0cm 4.5cm 0},clip,width=0.21\linewidth]{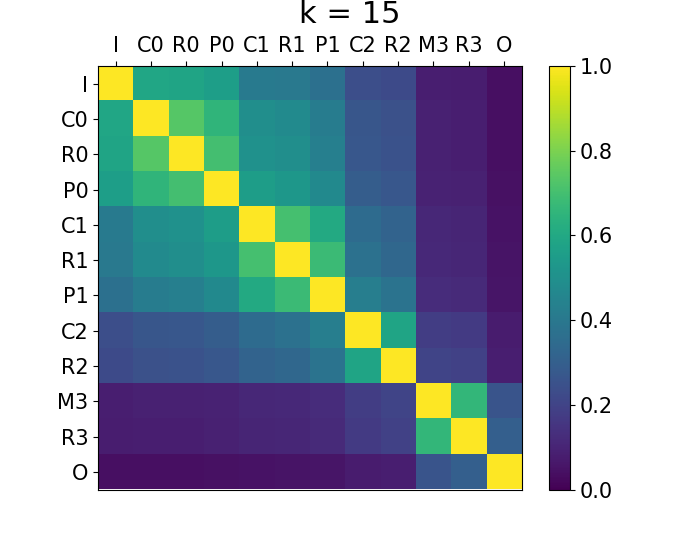} &
    \includegraphics[trim={1.5cm 1.0cm 4.5cm 0},clip,width=0.21\linewidth]{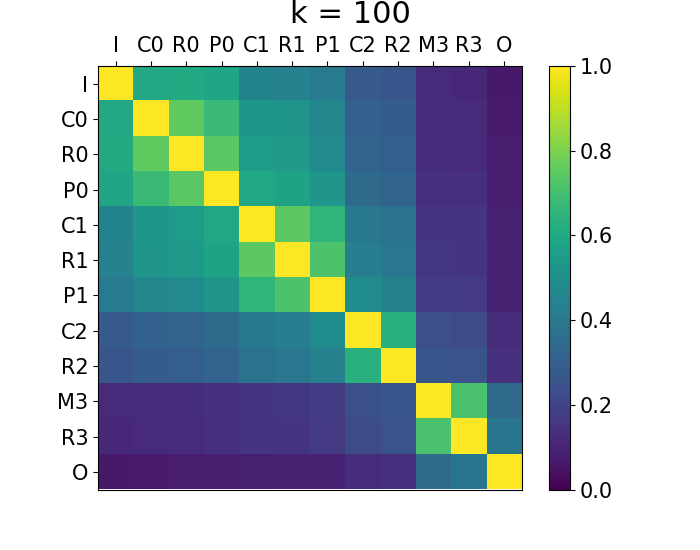} &
    \includegraphics[trim={1.5cm 1.0cm 4.5cm 0},clip,width=0.21\linewidth]{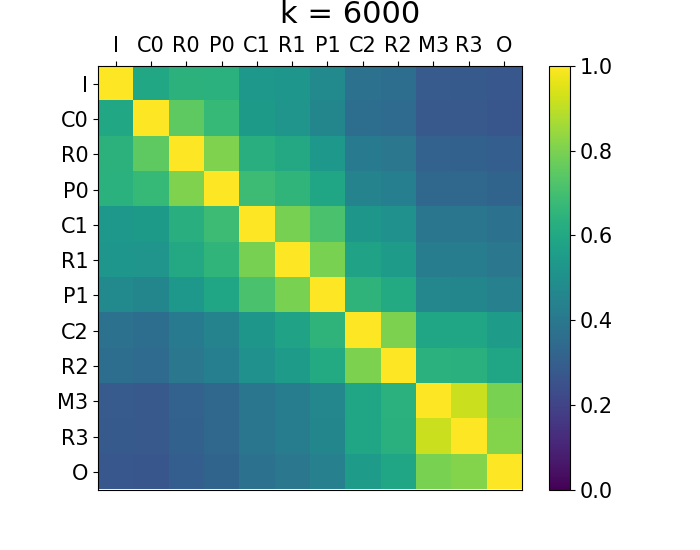} &
    \includegraphics[trim={1.5cm 1.0cm 2cm 0},clip,width=0.255\linewidth]{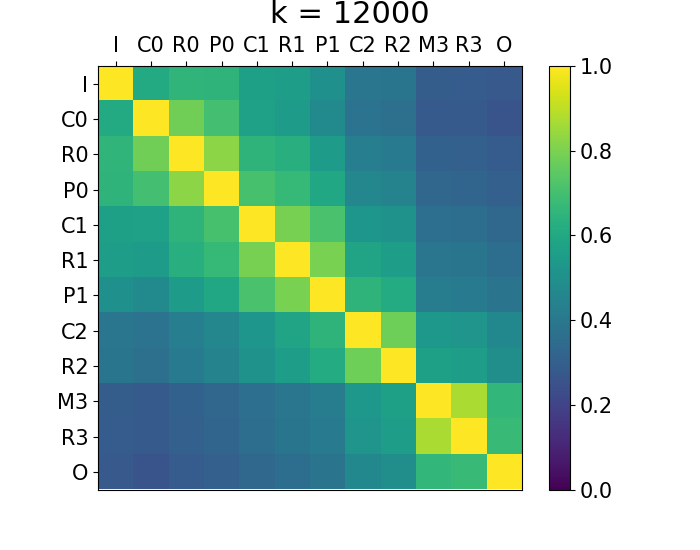}

    \end{tabular}
    \caption{\small The Nearest Neighbour Topological Similarity between layers in a LeNet-5 model trained on the MNIST dataset for a different number of $k$ nearest neighbours.}
    \label{fig:nnls}
\end{figure}

We apply the notion of Nearest Neighbour Topological Similarity (NNTS) to a LeNet-5~\cite{lecun1998gradient} architecture trained on the MNIST~\cite{lecun1998mnist} dataset to see how the local topological structure of a dataset changes across the model. Since LeNet-5 is a small architecture we break up what is normally considered a layer into their respective atomic operations before applying NNTS. We measure the NNTS between all pairs of operations in the LeNet-5 model. The results for NNTS analysis are shown in Figure~\ref{fig:nnls}. We show the NNTS matrix for four different values of $k$, 15, 100, 6000, and 12000. The table headers along the top and left indicate the operation with the LeNet-5 model. The operations are causally aligned moving from left to right along the top, and top to bottom on the left. I stands for the input layer, C for convolutional operations, R for ReLU activation, P for max-pooling, M for matrix multiplication, and O for output (note that O is also a matrix multiplication operation). The number following the operation identifier indicates the layer number.

Each of the NNTS matrices are symmetric about the diagonal. The diagonal in each plot all have values of 1 since layers are all self similar. Notice the block-like pattern staggered both vertically and horizontally in all four plots every time the layer number changes (e.g., moving from P0 to C1, or from R3 to O). This is a clear indication that sequences of operations which are normally considered \textit{layers} (within LeNet-5) are not arbitrary since internal representations within a layer are more similar to one another than to operations outside the layer. The inter-layer similarity pattern even persists when comparing the first operation in a layer to the first operation in the following layer. This observation could be used to study other standard layer designs (e.g., a layer designed with batchnorm) to determine if such designs follow the same inter-layer similarity pattern. Notice that the similarity between P1 and M3 is marginally smaller (about 0.05) when compared the similarity between R2 and M3. The marginal difference is a good indication that removing layer 2 will have little effect on network performance.

As the number of $k$ nearest neighbours increases from $k=15$ to $k=6000$ there is a gradual increase in the similarity between all layer pairs. The transition between from $k=6000$ to $k=12000$ sees a decrease in similarity, and most noticeably between the last couple of operations (bottom right hand corner), in the LeNet-5 model. This decrease is to be expected considering that MNIST has ten classes with approximately 6000 samples per class. Near the end of the network samples from the same class should be clustered near one another. At a $k=6000$ a sample's connections will mostly consist of all samples from within the class. Any operation performed on samples from the same class would likely have the same effect and thus not effect the inter-layer neighbour relations. But when $k=12000$ half a sample's neighbours will be from other classes. While operations are unlikely to effect intra-class neighbours, they can still effect inter-class neighbours, and thus resulting in the decrease in similarity from $k=6000$ to $k=12000$. It is expected that the inter-layer similarity converges to 1 as the number of connections approaches the number of samples in the dataset.

\section{Nearest Neighbour Topological Persistence}
\label{sec:nnnn_persistence}

Reducing the similarity between two layers to a single value provides a useful measure for high level measure for topological similarity. On the other hand, such reduction also removes most of the local inter-sample relationship information, thereby reducing one's ability to study the complex interactions between layers throughout a network. In this section we introduce an approach from which higher order analysis can be performed. Specially, we investigate when pairs of samples become neighbours in a DNN, properties of the pairs while they are neighbours, and when pairs of samples are no longer neighbours. We call such analysis Nearest Neighbour Topological Persistence (NNTP).

Consider a network where layers follow a sequential design $v_{in} > \cdots > v_a > \cdots > v_b > \cdots > v_{out} \in V$, where layer $v_{in}$ is the input layer and $v_{out}$ is the output layer of a DNN. Let $e_{ij}$ be an abstract un-directed connection between samples $(\textbf{x}_{i}, \textbf{x}_{j})$, and let $e^v_{ij}$ be the un-directed connection between samples $(\textbf{x}_{i}, \textbf{x}_{j})$ in layer $v$. Let $e^v_{ij} \in H_v$ iff either of the corresponding directional connections are in $H_v$, where $H_v$ is the NNG of layer $v$.

Let $e_{ij}$ be $\alpha{\text -}persistent$ between two layers $v_a$ and $v_b$ if there exists no more than $\alpha$ contiguous layers in the chain of layers $v_a > \cdots > v_b$ where $e^v_{ij} \notin H_v$ for all $v$'s within the chain of layers. $\alpha{\text -}persistent$ is a whole family of measures. In this work we only investigate $0{\text -}persistent$ sample pairs. $0{\text -}persistent$ can be interpreted as a measure for local network stability. If a connection persists across a series of layers then it is reasonable to assume that the pair is located in a local region of the dataset's manifold that share specific features. Analysing when samples are no longer neighbours may help illuminate what specific features a given network layer is detecting. When considering the entire dataset using this approach one can see the interactions between layers. For example, aspects of a network like connection-cancellation would become evident (i.e., if one layer moves a lot of samples near each other and a down stream layer moves those samples apart). By studying how layers interact with each other on a more granular level (when compared to scalar LS measures) one can tailor a DNN's design at both a macro-architecture resolution and a micro-architecture resolution.

We apply the notion of persistence to a LeNet-5 architecture trained on the MNIST dataset. We break the LeNet-5 model into the same atomic operations as done in the previous section. For $0{\text -}persistent$ analysis we count how many pair-wise nearest neighbour connections are $0{\text -}persistent$ between all pairs of layers in the LeNet-5 model. The results for $0{\text -}persistent$ analysis are shown in Table~\ref{tab:nnp_persistence_atomic_table}. The headers along the top and left indicate the operation with the LeNet-5 model. The operations are causally aligned when moving from left to right a long the top, and top to bottom on the left. We use $k=15$ for the number of neighbours each sample has.

\newcolumntype{"}{@{\hskip\tabcolsep\vrule width 2pt\hskip\tabcolsep}}
\begin{table*}
    \footnotesize
    \centering
    \caption[LeNet-5 $0{\text -}Persistent$ Matrix]{LeNet-5 $0{\text -}persistent$ matrix. Each operation pair $(v_{first}, v_{last})$ counts the number of $0{\text -}persistent$ pairwise samples (in the thousands) that start at operation $v_{first}$ and last appear at operation $v_{last}$. }
    \begin{tabular}{l c " c |ccc|ccc|cc|cc|c}
    \multicolumn{14}{c}{\centering\textbf{Layer of $0{\text -}persistent$ end}} \\
    \multirow{14}{*}{\rotatebox[origin=c]{90}{\centering\textbf{Layer of $0{\text -}persistent$ beginning}}} & & I & C0 & R0 & P0 & C1 & R1 & P1 & C2 & R2 & M3 & R3 & O \\
    \cmidrule[2pt](lr){2-14}
    & I & 455 & 122 & 6.99 & 5.95 & 2.51 & 3.02 & 36.0 & 0.99 & 0.59 & 3.94 & 0.32 & 28.7 \\
    \cmidrule(lr){2-14}
    & C0 &  & 427 & 2.94 & 5.30 & 0.65 & 0.52 & 2.27 & 1.23 & 0.06 & 0.18 & 0.01 & 0.62 \\
    & R0 &  &  & 550 & 2.72 & 0.57 & 0.61 & 10.2 & 0.13 & 0.02 & 0.09 & 0.01 & 0.46 \\
    & P0 &  &  &  & 494 & 19.0 & 22.1 & 6.08 & 0.54 & 0.35 & 2.50 & 0.42 & 0.53 \\
    \cmidrule(lr){2-14}
    & C1 &  &  &  &  & 110 & 114 & 286 & 1.71 & 0.17 & 0.82 & 0.10 & 2.83 \\
    & R1 &  &  &  &  &  & 67.2 & 57.7 & 1.33 & 0.23 & 0.29 & 0.04 & 0.39 \\
    & P1 &  &  &  &  &  &  & 205 & 4.53 & 2.54 & 0.57 & 0.12 & 0.55 \\
    \cmidrule(lr){2-14}
    & C2 &  &  &  &  &  &  &  & 528 & 36.3 & 8.79 & 0.57 & 0.46 \\
    & R2 &  &  &  &  &  &  &  &  & 291 & 104 & 10.4 & 168 \\
    \cmidrule(lr){2-14}
    & M3 &  &  &  &  &  &  &  &  &  & 170 & 19.1 & 141 \\
    & R3 &  &  &  &  &  &  &  &  &  &  & 149 & 128 \\
    \cmidrule(lr){2-14}
    & O &  &  &  &  &  &  &  &  &  &  &  & 180 \\

    \end{tabular}
    \label{tab:nnp_persistence_atomic_table}
\end{table*}

Notice the large number of connections present along the diagonal. These connections only sequentially exist for one layer (note that they may reappear in other layers). Let connections along the diagonal be called transient connections. Many layers in the LeNet-5 model have a plurality of their pairwise sample connections existing as transient connections, with the first layer (i.e. layer 0) being especially transient heavy. This may indicate that the first couple of operations are mainly responsible for placing the samples in approximately their final location in the data's manifold for classification, with the rest of the layers being responsible for fine tuning.

Another interesting observation is the number of connections present in the top right layer pair (I,O). These connections persist throughout all operations in the LeNet-5 model, indicating that they are likely to be true neighbours on the data's intrinsic manifold. Studying the relationship between such neighbours would be useful in a number of areas including building more robust datasets, tracking clusters of strongly persistent neighbours (i.e., connections that are persistent across many layers), and training a model on a reduced number of samples.

From this matrix one can see that C2 and R2 seem to have little effect on the data manifold as they largely add persistent connections while allowing most other connections to pass though. For applications like layer reduction, C2 and R2 are potentially strong candidates for layer removal, and even more so considering that C2 has the largest number of parameters when compared to the other convolutional operations. One anomaly with C2 is that it largely kills connections created by C1 as indicated by the operation pair (C1, P1) of 286000. Notice that 286000 is by far the largest non-transient group of connections in Table~\ref{tab:nnp_persistence_atomic_table}. In a sense, C2 is undoing the alterations to the data manifold made by C1. In addition, such a relationship does not exist between C0 and C1, or C0 and C2. Further research is required to understand such behavior.

\section{Conclusion and Future Work}
\label{sec:pnn_summary}

We propose two complementary data centric analytic methods for studying the complex dynamics of a dataset's manifold as it traverses through a DNN using a set of nearest neighbour graphs. The first proposed approach, Nearest Neighbour Topological Similarity, measures the local similarity between two NNGs, while the second proposed approach, Nearest Neighbour Topological Persistence, captures the complex local interactions between layers. We demonstrate that both these approaches have the potential for providing a better understanding of interactions between layers on a local topological level, and how such insights can be used to build better DNNs. Future directions of research include, but not limited to, using the proposed approach to study local clusters of data throughout a DNN, studying how a family of operations (e.g., activation functions) effects local characteristics of a dataset's manifold, and measuring how a manifold changes throughout training a DNN.

\medskip
\small

\bibliographystyle{ieeetr}
\bibliography{refs}

\begin{thebibliography}{1}

\bibitem{lecun2015deep}
Y.~LeCun, Y.~Bengio, and G.~Hinton, ``Deep learning,'' {\em nature}, vol.~521,
  no.~7553, pp.~436--444, 2015.

\bibitem{facco2017estimating}
E.~Facco, M.~d’Errico, A.~Rodriguez, and A.~Laio, ``Estimating the intrinsic
  dimension of datasets by a minimal neighborhood information,'' {\em
  Scientific Reports}, 2017.

\bibitem{ansuini2019intrinsic}
A.~Ansuini, A.~Laio, J.~H. Macke, and D.~Zoccolan, ``Intrinsic dimension of
  data representations in deep neural networks,'' in {\em Advances in Neural
  Information Processing Systems (NIPS)}, 2019.

\bibitem{gretton2005measuring}
A.~Gretton, O.~Bousquet, A.~Smola, and B.~Sch{\"o}lkopf, ``Measuring
  statistical dependence with hilbert-schmidt norms,'' in {\em International
  Conference on Algorithmic Learning Theory}, Springer, 2005.

\bibitem{hardoon2004canonical}
D.~R. Hardoon, S.~Szedmak, and J.~Shawe-Taylor, ``Canonical correlation
  analysis: An overview with application to learning methods,'' {\em Neural
  Computation}, 2004.

\bibitem{raghu2017svcca}
M.~Raghu, J.~Gilmer, J.~Yosinski, and J.~Sohl-Dickstein, ``Svcca: Singular
  vector canonical correlation analysis for deep learning dynamics and
  interpretability,'' in {\em Advances in Neural Information Processing Systems
  (NIPS)}, 2017.

\bibitem{kornblith2019similarity}
S.~Kornblith, M.~Norouzi, H.~Lee, and G.~Hinton, ``Similarity of neural network
  representations revisited,'' {\em arXiv preprint arXiv:1905.00414}, 2019.

\bibitem{lecun1998gradient}
Y.~LeCun, L.~Bottou, Y.~Bengio, and P.~Haffner, ``Gradient-based learning
  applied to document recognition,'' {\em Proceedings of the IEEE}, 1998.

\bibitem{lecun1998mnist}
Y.~LeCun, ``The mnist database of handwritten digits,'' {\em http://yann.
  lecun. com/exdb/mnist/}, 1998.

\end{thebibliography}

\newpage

\normalsize
\appendix
\section{Nearest Neighbour Graph}
\label{appendix:knn_graph}

Let $\textbf{x}_i \in \textbf{X}$ be a set of input samples of shape $n \times d$, and let $G=(V, E, W)$, represent a DNN, where $V=\{v\}$ is a set of sub-functions, $E=\{e\}$ is a set of edges that represent the sub-function's i/o relationships, and $W=\{w\}$ is a set of weights that parameterize the sub-functions. Let the output of some sub-function $v_v$ for the $\textbf{x}_i$ sample be defined as $\textbf{y}_{vi} = v_{v}(\textbf{x}_i; G_v)$, where $G_v \subseteq G$ contains all required sub-functions, edges, and weights to calculate $\textbf{y}_{vi}$.

The main idea behind our approach is to use a graph of neighbours to capture the local structure between samples within a layer. More formally, let $\textbf{Y}_{v} = v_{v}(\textbf{X}; G_v)$ be a one-to-one mapping of samples from the input space to the space of layer $v_v$ of a DNN. For a given layer $v_v$ with a set of outputs $\textbf{Y}_{v}$, let $H_v = (\textbf{Y}_v, D_v)$ be the graph of neighbours for layer $v_v$, where $\textbf{Y}_{v}$ are the vertices of the graph, and $D_v$ are the edges between two given samples $\textbf{y}_{vi},\textbf{y}_{vj} \in \textbf{Y}_{v}$. Let $K_{vi} \subseteq \textbf{Y}_{v}$ be an ordered set of nearest neighbours of sample $\textbf{y}_{vi}$.

The goal of the graph is to represent localized information from the samples. As such, a metric for measuring distance between two samples in a given layer is required. In general there are two common methods used. The first approach uses a distance threshold to find all samples $\textbf{y}_{vj} \in K_{vi}$ that are within some fixed radius $r_v$ of sample $\textbf{y}_{vi}$, where $r_v$ is constant for the entire graph. Note that each set $K_{vi}$ for a single layer $v_v$ can contain a variable number of neighbours. The second approach uses a variable radius but a fixed number of samples $k$ in $K_{vi}$ for each sample $\textbf{y}_{vi}$. Such an approach is called a $k$ nearest neighbour ($k$-nn) graph. For this work a $k$-nn based approach is used to ensure that each sample $\textbf{y}_{vj} \in \textbf{Y}_{v}$ has a neighbour (i.e., $|K_{vi}| > 0$). Note that it would be possible to find the smallest radius such that every sample has at least one neighbour, but this would also allow for samples to be connected to the entire graph (e.g., when there is one extreme outlier).

To build a $k$-nn graph one must choose if connections are directed or un-directed, what distance metric to use, and the number of neighbours. One of the features a distance metric requires is that the metric be commutative (i.e., $\langle x,y \rangle = \langle y,x \rangle $). From a $k$-nn graph's perspective this requires that connections between samples be un-directed. That is, if sample $\textbf{y}_{vi}$ is a neighbour of $\textbf{y}_{vj}$, then $\textbf{y}_{vj}$ must also be a neighbour of $\textbf{y}_{vi}$. However, the un-directed nature of connections would require a loosening of the fixed number of neighbours inherent to $k$-nn graphs as a $k$-nn graph with $k$ un-directed edges per sample may not exist.

One way to loosen the neighbourhood criteria is to perform an intersection where by two samples are un-directed neighbours iff both samples are directed neighbours of each other; this effectively sets an upper bound to the number of neighbours to $k$. Such an approach undermines the choice of a $k$-nn graph in that some samples might not have neighbours. Another way to solve the issue is to perform a union where by two samples are un-directed neighbours iff either sample is a directed neighbour of one another; effectively setting $k$ as the lower bound to the number of neighbours. This approach can result in some samples having orders of magnitude more neighbours then other samples. A third option to loosen the neighbourhood criteria is to just use directed edges, thereby ensuring every sample has the same number of neighbours. In this proposal directed edges are used for nearest neighbour graph (NNG) construction, other graph representations will be studied in the future work. For this work a euclidean based distance metric is used.

\end{document}